# Online trajectory recovery from offline handwritten Japanese kanji characters

Hung Tuan NGUYEN[†], Tsubasa NAKAMURA[‡], Cuong Tuan NGUYEN[‡] and Masaki NAKAGAWA[‡]

[†] Institute of Global Innovation Research

[‡] Department of Computer and Information Sciences

Tokyo University of Agriculture and Technology, Tokyo, 184-8588 Japan

E-mail: ntuanhung@gmail.com, s184843q@st.go.tuat.ac.jp, ntcuong2103@gmail.com and nakagawa@tuat.ac.jp

**Abstract** In general, it is straightforward to render an offline handwriting image from an online handwriting pattern. However, it is challenging to reconstruct an online handwriting pattern given an offline handwriting image, especially for multiple-stroke character as Japanese kanji. The multiple-stroke character requires not only point coordinates but also stroke orders whose difficulty is exponential growth by the number of strokes. Besides, several crossed and touch points might increase the difficulty of the recovered task. We propose a deep neural network-based method to solve the recovered task using a large online handwriting database. Our proposed model has two main components: Convolutional Neural Network-based encoder and Long Short-Term Memory Network-based decoder with an attention layer. The encoder focuses on feature extraction while the decoder refers to the extracted features and generates the time-sequences of coordinates. We also demonstrate the effect of the attention layer to guide the decoder during the reconstruction. We evaluate the performance of the proposed method by both visual verification and handwritten character recognition. Although the visual verification reveals some problems, the recognition experiments demonstrate the effect of trajectory recovery in improving the accuracy of offline handwritten character recognition when online recognition for the recovered trajectories are combined.

## 1. Introduction

For handwritten character recognition, there are two types of patterns: offline patterns of images from scanners and cameras and online patterns of touch/pen-tip trajectories from touch/pen-based devices. An online pattern consists of time-sequences of (x, y) coordinates so that an offline pattern can be easily reconstructed from the online pattern, but not vice versa. Although online patterns provide more productive features for classification, their variation in writing direction, writing speed, and stroke writing order was a challenge for online handwriting recognition so that offline features were added [1] or offline recognizers were combined [2]. On the other hand, offline patterns are free from the variations as mentioned above but lack some distinctive features. If online trajectories are reconstructed from offline patterns, it will improve offline handwritten character recognition. When experts decode damaged characters in historical documents, they often imagine their trajectory hypotheses to decode them. Therefore, the trajectory recovery may also be useful for them.

So far, offline handwriting recognizers have not been combined with online handwriting recognizers since online patterns are difficult to reconstruct from offline patterns [3]. There have been many studies on online handwriting recovery from offline patterns in Alphabet, Latin, Chinese scripts, as well as mathematic expressions [4]–[8]. Japanese kanji characters of Chinese origin are generally composed of many strokes (up to 30 strokes), as shown in Fig. 1 so that to recover their order is a challenge, among other languages.

Fortunately, there are some large online Japanese handwriting databases [9] which cover almost daily-used Japanese characters collected from hundreds of writers. In addition, online patterns can be easily converted to offline patterns without ambiguities. It is well known that offline patterns converted from online patterns are useful for training and evaluating offline recognizers [10], [11]. Therefore, we first convert online patterns to offline patterns and then try to recover online patterns from the above converted offline patterns. We can inspect the trajectory recovery because we have original online patterns. Moreover, we can evaluate the quality of trajectory recovery by online and offline recognitions for the original online patterns, the converted offline patterns, and the reconstructed online patterns from the converted offline patterns. Their combination may demonstrate interesting insights.

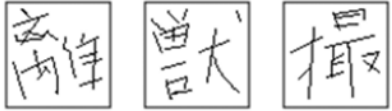

Fig. 1. Examples of Japanese characters written by multiple strokes.

We propose a data-driven method for trajectory recovery based on Deep Neural Networks (DNNs). A stimulating work was presented by Kumarbhunia et al. for Indian character/word patterns composed of single strokes [12]. The trajectory recovery for characters composed of many strokes, such as Japanese kanji and Chinese characters, seems to have a more significant potential benefit because stroke order variations and resultant pattern variations are large. According to the characteristics of writing kanji characters, we propose a specific neural network architecture to extract local features as well as maintain the 2-D spatial relations among all strokes, forming a kanji character pattern. We also demonstrate the necessity of the attention mechanism to focus on local context while reconstructing online trajectories sequentially.

The rest of the paper is organized as follows. In Section II, we present related work of online trajectory recovery with both traditional and deep learning-based methods. The details of our proposed method are declared in Section III. For training and evaluating the proposed method, we express them in Section IV and V, respectively. Finally, we give conclusions and discussions in Section VI.

## 2. Related works

In general, an online trajectory recovery method from an offline pattern usually has two components: a local feature extractor and a global generator. The local feature extractor extracts local context as well as detects junctions, starting and ending points, which provides local-level information to reconstruct the respective online trajectory. The global generator refers to the above local features for reconstructing based on some prior knowledge or heuristic rules. Jaeger et al. proposed an approach based on the shortest path problem in graph theory by expressing skeleton images of offline handwritten characters as graphs [8]. They employed the angles of intersecting lines for guiding global optimization. However, their method had the problem of combinatorial explosion on complex images that contain several ambiguous zones, such as junctions. Kato and Yasuhara proposed a similar method, which solved the problem of combinatorial explosion using heuristic rules to select a specific path in ambiguous zones [7]. These methods were proposed to recover online trajectories from single-stroke character images. However, they are difficult to employ for multiple-stroke character images due to the complexity of the stroke order.

Qiao and Yasuhara presented a method to recover online trajectories from multiple-stroke character images based on a bidirectional search to match the template strokes [4]. Their bidirectional search algorithm used dynamic programming to find the best matching the stroke order, which efficiently reduced the searching space. Cao et al. proposed an online trajectory recovery approach for Chinese characters, which are multiple-stroke characters [6]. To avoid a large number of character templates in the Chinese language, their approach consisted of three hierarchical levels to represent a Chinese character as components, subcomponents, and strokes, respectively. A character was analyzed to components, and each component was factorized to subcomponents in turn by four decomposing operators. Next, the relations between subcomponents were computed by four heuristic rules. Finally, the writing order of strokes was determined based on crossing stroke pairs. Another attempt to recovery online trajectories of multiple-stroke characters was proposed for handwritten mathematical symbols based on combinatorial reconstruction with heuristic search [5]. Note that the heuristic rules require a considerable effort to develop for a specific language, which is infeasible to cover the whole character set of Chinese or Japanese languages.

Inspired by the success of DNNs on sequence generation and image captioning, some studies employed DNNs for recovering an online trajectory from an offline image. Recently, an encoder-decoder model [12] has been proposed to recover an online trajectory from an offline handwritten character image, which is applied for Indian scripts with single-stroke characters/words. Its encoder consists of a Convolutional Neural Network (CNN) to extract low-level local features from an offline image, and a Long Short-Term Memory (LSTM) Network to compute high-level features (encoded features) from the low-level features. An LSTM-based decoder uses the encoded features to generate a sequence of online pen/touch points, forming the offline image. This research demonstrates the ability of DNNs to reconstruct an online trajectory from a handwritten character image. However, it is developed to recover single-stroke characters so that it is not applicable to Japanese kanji characters. Moreover, the proposed method is trained and evaluated based on low-level features of (x, y) coordinates, so that it limits the robustness of the decoder to deal with distortions and variations of handwriting.

## 3. Proposed method

Note that there are three main challenges for reconstructing online trajectories from Japanese kanji characters. A first challenge is a large number of strokes with a complex layout. The second one is the writing order of multiple strokes. The last one is a huge number of kanji characters in the Japanese language. Following the success of DNNs for sequence generation and online trajectory recovery by the encoder-decoder model [12], we propose a method also based on the encoder-decoder, but we incorporate an attention mechanism and GMM for the above-mentioned challenges.

In the case of a single-stroke character, the encoder with the LSTM network is appropriate to represent a 2-D image into a 1-D sequence of features since the relations among pen/touch points are mainly along the horizontal dimension. In contrast, Japanese kanji characters with multiple strokes have many relations along both horizontal and vertical dimensions, which should be preserved for reconstructing online trajectories. Thus, we eliminate the LSTM network from our encoder but insert the attention mechanism into our decoder.

As mentioned above, an encoder-decoder based network is trainable by the x- and y- coordinates output but easy to overfit with some specific writing styles in training samples, which may have difficulty for new writers. To improve the robustness of our proposed network, we assume our network output as a sequence of GMM parameters instead of a sequence of x- and y-coordinates.

### 3.1. Deep Neural Network-based encoder decoder

As proposed in [1], the encoder-decoder model provides a good mechanism to sequentially generate a new prediction based on the encoded features and previous predictions. It achieved high performances on different tasks such as machine translation [2], image captioning [3], and even handwritten mathematic expression recognition [4], where the spatial context is complex. Fig. 2 shows our proposed encoder-decoder model for online trajectory recovery from handwritten character images.

### 3.1.1. CNN-based encoder

Fig. 2 shows the structure of our CNN-based encoder on its left side. It is designed to extract local features from a 64×64 binary image. It consists of four Convolutional layers (CONV) with a kernel size of 3×3 and depth of 8, 32, 128, 256 (number of feature maps), respectively. Following the second and the third CONV, Batch Normalization layers (Batch Norm) is applied to increase the learning speed [5] when training with large batch size. The activation function for CONV is the Leaky Rectified Linear Unit (ReLU) function [6], which is a function with a small slope at $x<0$ of the ReLU function. Moreover, Max Pooling layers with a kernel size of 2×2 are employed to maintain the spatial context even in higher levels while Kumarbhunia et al. [7] used a kernel size of 1×2 to collect the information in the horizontal dimension of an image. The CNN-based encoder is composed of a small number of layers to avoid the overfitting problem since handwritten character images are more straightforward than general images [8].

### 3.1.2. LSTM-based decoder

The LSTM-based decoder is the state-of-the-art method for generating a sequence of predictions from encoded features. During the training stage, it loops through $T$ timesteps, which is equal to the most extended online trajectory target in a mini-batch. However, $T$ is fixed during the evaluating stage when the online trajectory target is not available. The online trajectory target is a sequence of online pen/touch points. Each online point is represented by ($x$, $y$) coordinates as well as its state represented by a one-hot vector, ($ps_0$, $ps_1$, $ps_2$), where $ps_0$ indicates the pen/touch-down state, $ps_1$ dose the pen/touch-up state and $ps_2$ expresses the end of the online trajectory. This representation of an online trajectory is referred from the sketch generating model [9], which is appropriate to compute loss functions during training.

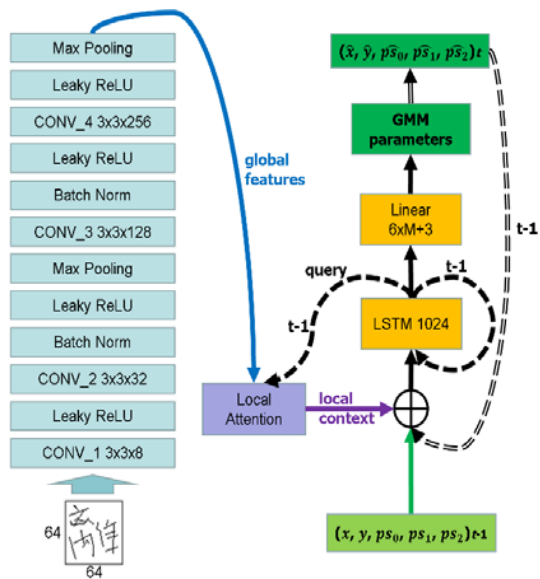

Fig. 2. Proposed encoder-decoder network with an attention layer to reconstruct an online trajectory from an offline image.

Fig. 2 shows the structure of our LSTM-based decoder on its right side at the $t$-th decoding timestep. The LSTM decoder receives the input vector $X_t$ as well as the recurrent vector of LSTM. The input vector consists of the local context by the attention layer, whose details are presented in the next section, and an online point.

For the training stage, we use the teacher forcing technique [10], so that the $t$-th online point in the input sequence is the $(t-1)$-th online point in the target sequence, $(x, y, ps_0, ps_1, ps_2)_{t-1}$ as represented by the green arrow in Fig. 2, and in Eq. (1):

$$X_t = \text{local\_context}_t \oplus (x, y, ps_0, ps_1, ps_2)_{t-1} \quad (1)$$

For the evaluating stage, the $t$-th online point in the input sequence is the prediction by the LSTM-based decoder at the $(t-1)$-th decoding timestep, $(\hat{x}, \hat{y}, \widehat{ps_0}, \widehat{ps_1}, \widehat{ps_2})_{t-1}$ as represented by the double dash arrow in Fig. 2, and in Eq. (2):

$$X_t = \text{local\_context}_t \oplus (\hat{x}, \hat{y}, \widehat{ps_0}, \widehat{ps_1}, \widehat{ps_2})_{t-1} \quad (2)$$

For the first decoding timestep ($t=0$) in both the training and evaluating stages, the online point is assigned by a specific vector (0,0,0,1,0).

The LSTM-based decoder uses an LSTM layer with 1024 cells and a linear layer with 6x$M$+3 units, where $M$ is the number of Gaussian distributions, whose details will be described in the following section.

### 3.1.3. Attention layer

As presented above, local attention is integrated into the decoder since Japanese handwritten kanji character images have a complex layout of multiple strokes. It is introduced to guide the decoder to focus on a specific local region at every decoding timestep. In machine translation, the introduction of the attention mechanism has been proposed to reduce the degradation of translation performance in long sentences [2], [11]. In our case, the attention layer is modified to compute **local context** from the **global features** encoded by the CNN-based encoder and the LSTM output at previous decoding timestep (**query**). The global features have the shape of $H_{gf} \times W_{gf} \times D_{gf}$, where $H_{gf}, W_{gf}, D_{gf}$ are height, width, and depth of global features, respectively.

First, the local **attention map** is computed by applying linear transformations to the global features and query and then applying ReLU, followed by Softmax, as shown in Eq. (3):

$$\text{attention\_map}_t = Softmax\left(ReLU(Linear(\text{global\_features}) + Linear(\text{query}_t))\right) \quad (3)$$

where **query**$_t$ is the output of the LSTM decoder at the $t$-1 timestep. The attention map focuses on the position of the online point at the current decoding timestep in the global features since the query consists of the previous decoded output, which is different at every decoding timestep.

Next, the local context vector is computed by multiplying the attention map and the global features at $(i, j)$ in the global features and summing these $H_{gf} \times W_{gf}$ products as shown in Eq (4):

$$\text{local\_context}_t = \sum_{\substack{0 \le i < H_{gf} \\ 0 \le j < W_{gf}}} \text{attention\_map}[i,j]_t \times \text{global\_features}[i,j] \quad (4)$$

where $H_{gf}$ and $W_{gf}$ are the height and width of the global features. Moreover, the attention layer is built from the differentiable operators so that it is trainable by stochastic gradient descent together with the whole network.

### 3.2. Gaussian Mixture Model parameters as output

As presented in the decoder section, the online trajectory is represented by a sequence of vectors $X_t = (x, y, ps_0, ps_1, ps_2)_t$, where $t$ is the temporal order of the online points. Moreover, $x$ and $y$ could be expressed as a probability distribution obtained by superimposing $M$ normal distributions using the Gaussian Mixture Model. The probability distribution $p(x,y)$ is calculated, as shown in Eq. (5):

$$p(x,y) = \sum_{m=1}^{M} \pi_m N(x,y | \mu_{x,m}, \mu_{y,m}, \sigma_{x,m}, \sigma_{y,m}, \rho_{xy,m}) \quad (5)$$

The sum of mixture weights is 1, as shown in Eq. (6):

$$\sum_{m=1}^{M} \pi_m = 1 \quad (6)$$

$N(x,y | \mu_{x,m}, \mu_{y,m}, \sigma_{x,m}, \sigma_{y,m}, \rho_{xy,m})$ represents a bivariate normal distribution.

For the $m$-th Gaussian distribution, $\mu_{x,m}, \mu_{y,m}$ are the mean values, $\sigma_{x,m}, \sigma_{y,m}$ are the standard deviation values, $\rho_{xy,m}$ is the correlation coefficient and $\pi_m$ is the mixture weight. Hence, the number of GMM parameters is 6x$M$, where $M$ is the number of Gaussian distributions. Moreover, every online point has its state represented by the vector of ($ps_0, ps_1, ps_2$). Therefore, the decoding output at each timestep consists of $6 \times M + 3$.

In order to compute the reconstruction loss from GMM parameters, we use Eq. (7) and Eq. (8):

$$L_s = -\frac{1}{N_p}\sum_{t=1}^{N_p} log(p(x_t, y_t)) \qquad (7)$$

$$L_p = -\frac{1}{T}\sum_{t=1}^{T}\left(\sum_{k=0}^{2} ps_{k,t} log(\widehat{ps}_{k,t})\right) \qquad (8)$$

where $N_p$ is the number of online points of the target trajectory. Note that the loss $L_p$ is computed at all timesteps, while $L_s$ is only computed depending on the length of the online trajectory ($N_p$). The general loss $L$ is the sum of $L_s$ and $L_p$. It is used to optimize the encoder-decoder network parameters by the stochastic gradient descent algorithm.

## 4. Training stage

This section expresses the details of data preparation and configurations to train our proposed network. Since there is no Japanese handwriting database collecting both online and offline patterns at the same time, we generate offline handwritten character images from online handwritten character patterns. Although the rendered images are not real offline patterns, they are useful to train and evaluate the performance of trajectory recovery, as suggested by [7].

### 4.1. Data preparation

A part of an online handwritten character pattern database: nakayosi_t-98-09 collected in the display coordinates from 163 writers is used as training data [12]. A part of another database: kuchibue_d-96-02 collected in the tablet coordinates from different 120 writers is used as testing data [12]. Using the different databases collected in different coordinates and from different writers, we could demonstrate the performance of our proposed trajectory recovery method in practical cases. In this experiment, training and evaluating are performed on kanji characters composed of more than 15 strokes in order to examine the ability of restoring stroke order information for complex kanji characters. In the training dataset, the number of character images is 120,005 from 1,324 categories while there are 60,293 samples from 713 categories in the testing dataset.

For normalizing the online patterns collected indifferent coordinates, we rescale every single online character pattern to the size of 64×64. Next, we apply the digitized point reduction algorithm [13], which shortens the online trajectory by omitting the redundant points. Moreover, the shorter online trajectories are more appropriate to train using the gradient descent algorithm. From the preprocessed online patterns, the offline handwritten character images are rendered by drawing the lines connecting online points on a bitmap. As suggested from [9], the online trajectory target for training should be padded by a vector, (0,0,0,0,1), following the longest online trajectory in a mini-batch so that the unbalance among three pen states is reduced. Moreover, a dummy vector (0,0,0,1,0) is inserted into the beginning of the trajectory, as mentioned in the decoder section.

### 4.2. Configurations for training network

In the following experiments, we employ GMM with 20 Gaussian distributions (or $M$=20). Thus, the linear layer after the LSTM layer has 20 x 6 + 3 = 123 units. In order to train the proposed network, we employ the Adam algorithm for optimizing parameters with a learning rate $10^{-3}$ [14]. The maximum number of epochs for training is fixed at 100. While doing experiments, however, we found that the model was usually overfitting to the training set after the 20-th epoch. For boosting the training stage, we employ mini-batch training with a batch size of 64 due to the limit of GPU memory. Our work is implemented using the PyTorch framework for automatically compute gradients and update weights [15].

## 5. Evaluating stage

### 5.1. Evaluation methods for online trajectory recovery

Besides many online trajectory recovery methods, many evaluation approaches have been proposed for measuring the quality of reconstructed online trajectories from offline patterns, as summarized in the survey [16]. The most straightforward approach is a visual evaluation, which is appropriate only in the case of a small number of testing samples. However, it is useful to analyze the erroneous cases so that the recovery method could be investigated to improve. Other elastic matching approaches, such as dynamic time warping (DTW) or Hidden Markov Models, match a recovered trajectory with its corresponding ground truth [17], [18]. These approaches, however, do not provide a measurement on the quality of reconstructed trajectories in practical situations.

In this paper, we present two evaluation approaches for our proposed trajectory recovery method. The first approach is visual verification, and the second one is recognition performance evaluation by handwritten character recognition. In the first approach, we show some examples of both successful and unsuccessful samples, which are useful to consider how to revise the method in further researches. The second approach points out a quantitative measurement to determine whether our proposed method can be used to improve offline handwritten character recognition. We reuse the pre-trained online handwritten character recognizer by Nguyen et al. [19] because it was trained on the same databases. This recognizer is robust with various writing styles while it is still sensitive to wrong stroke orders, missing strokes, or spurious strokes.

### 5.2. Evaluation by visual verification

This section presents visual verification on online trajectory recovery with and without the attention layer for some typical samples. In order to show the comparison, we also prepared and trained the same network architecture without the attention layer. Although movies are the best media to show online trajectory recoveries, we can only use static figures to show them in this paper so that we color the original strokes in black and the recovered strokes in red in the following figures.

Fig. 3 shows three successful recovered examples from the decoders with and without attention. Although the network without attention could reconstruct the online trajectories to some extent, they do not match the pixels in offline handwritten character images. The reason is that the encoder-decoder network extracts similar CNN-based features from different samples belonging to the same character category. Although the decoder has enough information to generate a sequence of online points, the reconstructed online trajectory is the same as some samples in the training dataset due to the lack of the attention layer. Thus, the attention layer is essential to avoid overfitting as well as to adapt to real samples.

Fig. 4 presents three samples that are successfully recovered with attention while not without attention. The first and second samples show samples correctly reconstructed with attention but mis-reconstructed without attention. The reconstructed trajectories without attention are different characters from the target characters. Without attention, the decoder refers to the whole encoded features, which might be similar to encoded features from other characters. In the third sample, the decoder without attention is stuck during the decoding. It might happen when the encoded features do not match any training sample.

On the other hand, Fig. 5 shows unsuccessfully recovered samples by both with and without attention. While the network without attention generates incorrect online trajectories, the network with the attention layer does not complete the decoding phase. In our proposed network, the local context at every decoding timestep is calculated by the attention layer based on both global features and previous decoded context. Thus, if the attention layer does not work properly, the outputs are unlikely reconstructed correctly.

To analyze the working of attention layer in both success and unsuccess cases, Fig. 6 shows the attention maps and reconstructed trajectories at specific decoding timesteps denoted as *t*. At each decoding timestep, an offline image with its attention map (brighter pixels represent for higher attention probabilities) and the reconstructed trajectory until the *t*-th decoding timestep are shown.

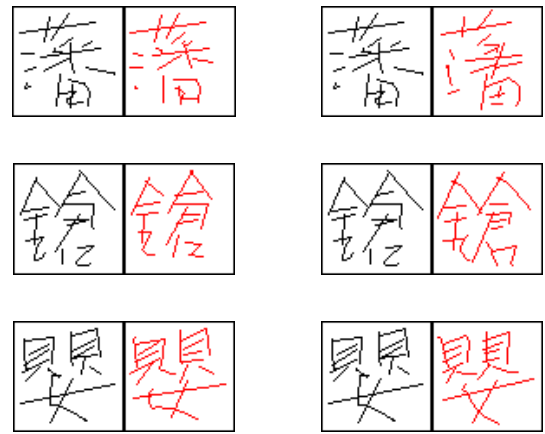

(a) With attention layer.    (b) Without attention layer.

Fig. 3.    Successfully recovered examples.

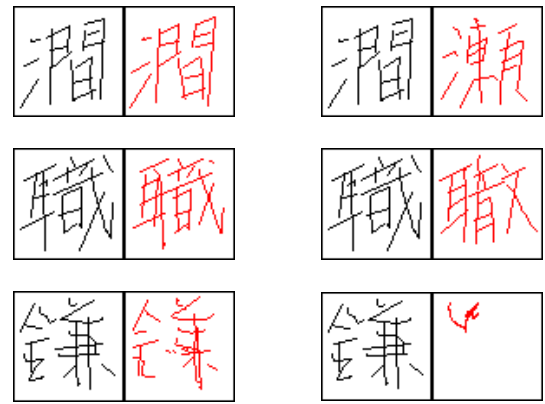

(a) With attention layer.    (b) Without attention layer.

Fig. 4.    Successfully recovered examples with attention.

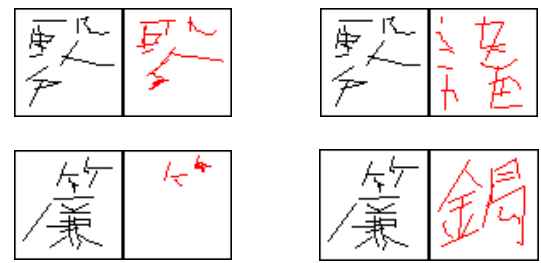

(a) With attention layer.    (b) Without attention layer.

Fig. 5.    Unsuccessfully recovered examples.

In Fig.6 (a), the attention maps and the recovered online trajectories roughly match the input image. However, the 7th stroke of this sample is extremely slow, which takes 8 steps to complete (from *t*=8 to *t*=15), which is unnatural compared to human writing. This unnatural online trajectory might cause the difficulty to predict the next online points and affect online handwritten character recognition because the online recognizer was trained using samples written by humans with natural writing speed.

In Fig. 6 (b), the attention maps stuck around the middle region of the offline image so that the online trajectory is incorrectly predicted. Since this unsuccessfully recovered sample is not particularly complex compared to other successfully recovered samples, there is room for improvement in the network structure, especially the attention mechanism.

### 5.3. Evaluation by handwritten character recognizer

To evaluate the quality of reconstructed trajectories, we employed offline and online recognizers. The offline handwritten character recognizer is composed of 4 Convolutional layers with kernel sizes of 3x3 and depths of 100, 200, 300, 400, respectively. Each Convolutional layer is followed by a Max Pooling layer. After the fourth Max Pooling layer, a Fully Connected layer of 500 ReLU cells with a dropout rate of 0.25 is employed. Finally, a classification layer based on the Softmax activation function is used to compute the recognition probabilities of a converted offline pattern. This offline recognizer was trained by the converted offline images from the training dataset used in this research.

On the other hand, we reused the BLSTM online recognizer from [19] since it was trained on the same nakayosi database. However, the online recognizer was trained using the original online points without resizing and reducing the redundant points. In the following experiment, we employed a simple weighted fusion of recognition scores to produce the combined recognizer, as shown in Eq. (9):

$$P_{comb} = \gamma \, log(P_{off}) + (1 - \gamma) \, log(P_{on}) \qquad (9)$$

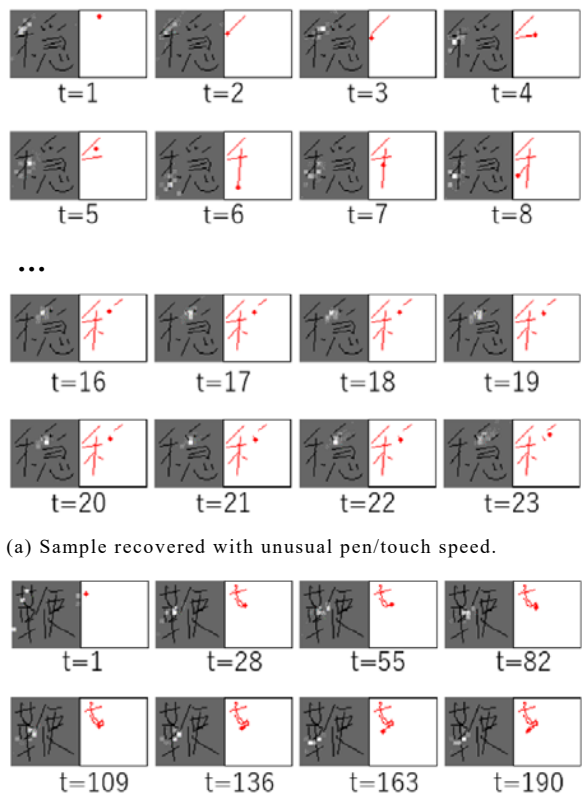

(a) Sample recovered with unusual pen/touch speed.

(b) Sample whose recovery was stuck.

Fig. 6. Attention maps and recovered trajectories by decoding timesteps.

where, $P_{off}$ and $P_{on}$ are the recognition scores from offline and online recognizers, respectively. The hyperparameter γ plays a role as a weight factor between offline and online recognizers.

We evaluated combined recognition with three different online patterns:

i) original online handwritten patterns from the testing dataset (kuchibue),

ii) preprocessed trajectories after resizing the original patterns to 64x64 and reducing their redundant points,

iii) reconstructed trajectories recovered by our method.

Fig. 7 shows the accuracies of the combined recognizer for the three different online patterns with γ from 0.4 to 1.0, where γ=1.0 denotes only the offline recognizer is used. The blue line represents the combined accuracy with the online recognizer for the original online patterns, the orange dash line for the preprocessed, and the green line for the reconstructed. The green line shows an improvement when using the reconstructed online trajectories with offline patterns in the combined recognizer. For example, the accuracy is improved from 97.47% to 98.13% when γ is 1.0 and 0.7, respectively. Because the reconstructed online trajectories are not complete and still need to be improved, they are challenging to contribute equally to offline images. Nevertheless, offline character recognition is helped by online recognition of reconstructed trajectories.

The blue line can be considered as the theoretical limit of the combined recognizer only when online trajectories are entirely reconstructed. In Fig. 7, there is a considerable gap between the blue and green lines, especially when γ becomes smaller. Because the training samples for our proposed method are the preprocessed online patterns, the recovered patterns should be similar to the preprocessed patterns rather than the original patterns, while the online recognizer was not trained by the same preprocessed patterns as reported in [19]. Hence, we need to confirm whether the preprocessing steps in this paper seriously affect the combined recognition performance. The differences between the blue line and the orange dash line are from 0.5 to around 1.4 percentage points, while the differences between the orange dash line and the green line are from 0.2 to more than 3.7 percentage points. Consequently, the network structure should be modified with a high priority, while the online recognizer can be retrained with the preprocessed online patterns.

Fig. 8 shows two examples that were misrecognized by the offline recognizer for their offline handwritten character images but correctly recognized by the combined recognizer for their offline and the reconstructed online character patterns. In Fig 8 (a) and (b), the kanji characters under the offline images colored in black are the recognition results by the offline recognizer, while those under the reconstructed online trajectories colored in red are recognition results by the combined recognizer. In the case in Fig. 8 (a), two kanji characters recognized by the offline recognizer and the combined recognizer look similar, especially when we only consider the offline image. However, their online trajectories are different, especially the number of strokes. Thus, the reconstructed online trajectory helps the combined recognizer not misrecognize it. In Fig. 8 (b), some strokes are written in touch, which causes misrecognition when using only the offline image. Due to the correctly reconstructed online trajectory with their separated strokes, the combined recognizer correctly recognizes it.

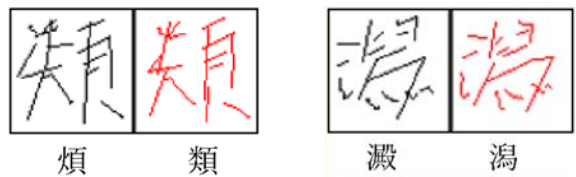

(a) Ambiguous offline image.  (b) Touching strokes.

Fig. 8. Two correctly recognized examples by the combined recognizer.

## 6. Conclusions and Discussions

In this paper, we presented a method for restoring a dynamic online trajectory from an offline handwritten character image. By integrating an attention layer into the encoder-decoder network, the quality of reconstructed online trajectories is enhanced for the Japanese kanji characters with many strokes. We evaluated the reconstructed online trajectories by visual verification and handwritten character recognizer. Using the reconstructed online trajectories, we succeeded in improving the recognition rate of offline handwritten characters. This is probably the first report that offline character recognition is helped by online recognition of reconstructed trajectories.

We will analyze the causes of failures in trajectory recovery and enhance it so that online recognition of reconstructed online trajectories and the combined recognition could be further improved. Then, we will test our method for kanji characters composed of smaller and middle numbers of strokes. Moreover, we will test our method to databases with both offline and online patterns, such as CASIA-OLHWDB and CASIA-HWDB [20].

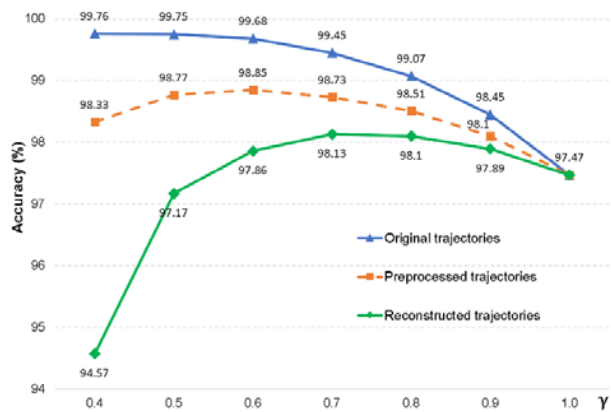

Fig. 7. Performance of online and offline recognizers with γ from 0.4 to 1.0.

## References

[1] K. Cho et al., "Learning Phrase Representations using RNN Encoder–Decoder for Statistical Machine Translation," in Proceedings of the 11th Conference on Empirical Methods in Natural Language Processing, 2014, pp. 1724–1734.

[2] D. Bahdanau, K. Cho, and Y. Bengio, "Neural Machine Translation By Jointly Learning To Align and Translate," in Proceedings of the 3rd International Conference on Learning Representations, 2015, pp. 1–15, doi: 10.1146/annurev.neuro.26.041002.131047.

[3] O. Vinyals, A. Toshev, S. Bengio, and D. Erhan, "Show and tell: A neural image caption generator," Proceedings of the IEEE Computer Society Conference on Computer Vision and Pattern Recognition, vol. 07-12-June, pp. 3156–3164, 2015, doi: 10.1109/CVPR.2015.7298935.

[4] J. Zhang et al., "Watch, attend and parse: An end-to-end neural network based approach to handwritten mathematical expression recognition," Pattern Recognition, vol. 71, pp. 196–206, Nov. 2017.

[5] S. Ioffe and C. Szegedy, "Batch Normalization: Accelerating Deep Network Training by Reducing Internal Covariate Shift," in Proceedings of the 32nd International Conference on Machine Learning, vol. 37, 2015, pp. 448–456, doi: 10.1007/s13398-014-0173-7.2.

[6] B. Xu, N. Wang, T. Chen, and M. Li, "Empirical Evaluation of Rectified Activations in Convolutional Network," CoRR, vol. abs/1505.0, 2015.

[7] A. Kumarbhunia et al., "Handwriting Trajectory Recovery using End-to-End Deep Encoder-Decoder Network," in Proceedings of the 24th International Conference on Pattern Recognition, 2018, pp. 3639–3644, doi: 10.1109/ICPR.2018.8546093.

[8] O. Russakovsky et al., "ImageNet Large Scale Visual Recognition Challenge," International Journal of Computer Vision, vol. 115, no. 3, pp. 211–252, Dec. 2015, doi: 10.1007/s11263-015-0816-y.

[9] D. Ha and D. Eck, "A Neural Representation of Sketch Drawings," in Proceedings of the 6th International Conference on Learning Representations, 2018, pp. 1–15.

[10] R. J. Williams and D. Zipser, "A Learning Algorithm for Continually Running Fully Recurrent Neural Networks," Neural Computation, vol. 1, no. 2, pp. 270–280, Jun. 1989, doi: 10.1162/neco.1989.1.2.270.

[11] T. Luong, H. Pham, and C. D. Manning, "Effective Approaches to Attention-based Neural Machine Translation," in Proceedings of the 12th Conference on Empirical Methods in Natural Language Processing, 2015, pp. 1412–1421, doi: 10.18653/v1/d15-1166.

[12] M. Nakagawa and K. Matsumoto, "Collection of on-line handwritten Japanese character pattern databases and their analyses," International Journal on Document Analysis and Recognition, vol. 7, no. 1, pp. 69–81, 2004, doi: 10.1007/s10032-004-0125-4.

[13] D. H. Douglas and T. K. Peucker, "Algorithms for the Reduction of the Number of Points Required to Represent a Digitized Line or its Caricature," Cartographica: The International Journal for Geographic Information and Geovisualization, vol. 10, no. 2, pp. 112–122, Dec. 1973, doi: 10.3138/fm57-6770-u75u-7727.

[14] D. P. Kingma and J. L. Ba, "Adam: a Method for Stochastic Optimization," in Proceedings of the 3rd International Conference on Learning Representations, 2015.

[15] A. Paszke et al., "PyTorch: An Imperative Style, High-Performance Deep Learning Library," in Advances in Neural Information Processing Systems 32, 2019, pp. 8024–8035.

[16] V. Nguyen and M. Blumenstein, "Techniques for static handwriting trajectory recovery: A survey," in Proceedings of the 9th IAPR International Workshop on Document Analysis Systems, 2010, pp. 463–470, doi: 10.1145/1815330.1815390.

[17] R. Niels and L. Vuurpijl, "Automatic Trajectory Extraction And Validation Of Scanned Handwritten Characters," in Proceedings of the 10th International Workshop on Frontiers in Handwriting Recognition, 2006, pp. 343–348.

[18] E. M. Nel, J. A. du Preez, and B. M. Herbst, "Verification of dynamic curves extracted from static handwritten scripts," Pattern Recognition, vol. 41, no. 12, pp. 3773–3785, Dec. 2008, doi: 10.1016/j.patcog.2008.05.005.

[19] H. T. Nguyen, C. T. Nguyen, and M. Nakagawa, "Online Japanese Handwriting Recognizers using Recurrent Neural Networks," in Proceedings of the 16th International Conference on Frontiers in Handwriting Recognition, 2018, pp. 435–440, doi: 10.1109/ICFHR-2018.2018.00082.

[20] C. L. Liu, F. Yin, D. H. Wang, and Q. F. Wang, "CASIA online and offline Chinese handwriting databases," in Proceedings of the 11th International Conference on Document Analysis and Recognition, 2011, pp. 37–41, doi: 10.1109/ICDAR.2011.17.